\documentclass[letterpaper]{article} 
\usepackage{aaai2026}  
\usepackage{times}  
\usepackage{helvet}  
\usepackage{courier}  
\usepackage[hyphens]{url}  
\usepackage{graphicx} 
\usepackage{natbib}  
\usepackage{caption} 
\usepackage{algorithm}
\usepackage{algorithmic}
\usepackage{subcaption}
\usepackage{amssymb}
\usepackage{booktabs}
\usepackage{amsmath}
\usepackage{multirow}

\usepackage{newfloat}
\usepackage{listings}
\DeclareCaptionStyle{ruled}{labelfont=normalfont,labelsep=colon,strut=off} 
\floatstyle{ruled}
\newfloat{listing}{tb}{lst}{}
\floatname{listing}{Listing}
\title{Breaking the Adversarial Robustness-Performance Trade-off in Text Classification via Manifold Purification}
\author {
    Chenhao Dang\textsuperscript{\rm 2},
    Jing Ma\textsuperscript{\rm 1}\thanks{Corresponding author.}
}
\affiliations {
    \textsuperscript{\rm 1}BRAIN, Renmin University of China\\
    \textsuperscript{\rm 2}China Electronics Technology Group Corporation 15th Research Institute\\
    dangchenhao@std.uestc.edu.cn, majingmady@ruc.edu.cn
}

\usepackage{bibentry}

\begin{document}

\maketitle

\begin{abstract}
A persistent challenge in text classification (TC) is that enhancing model robustness against adversarial attacks typically degrades performance on clean data. We argue that this challenge can be resolved by modeling the distribution of clean samples in the encoder’s embedding manifold. To this end, we propose the Manifold-Correcting Causal Flow ($MC^{2}F$), a two-module system that operates directly on sentence embeddings. A Stratified Riemannian Continuous Normalizing Flow (SR-CNF) learns the density of the clean data manifold. It identifies out-of-distribution embeddings, which are then corrected by a Geodesic Purification Solver. This solver projects adversarial points back onto the learned manifold via the shortest path, restoring a clean, semantically coherent representation. We conducted extensive evaluations on text classification (TC) across three datasets and multiple adversarial attacks. The results demonstrate that our method, $MC^{2}F$, not only establishes a new state-of-the-art in adversarial robustness but also fully preserves performance on clean data, even yielding modest gains in Accuracy.
\end{abstract}


\section{Introduction}

The remarkable success of Pre-trained Language Models (PLMs) like BERT across numerous Natural Language Processing (NLP) tasks~\cite{devlin2019bert} is shadowed by their pronounced vulnerability to adversarial attacks~\cite{jin2020bert, ren2019generating}. These attacks introduce minor, often semantically imperceptible, perturbations to input text, yet can trigger catastrophic prediction errors in the model, which becomes especially critical in text classification (TC).~\cite{vazquez2024survey,goyal2023survey, li2020bert,wang2021measure} In response, a wide array of defense mechanisms have been developed to enhance model robustness~\cite{gao2023dsrm, zheng2024subspace, asl2024robustsentembed, zhu2020freelb}. However, their success is often marred by a critical challenge: improving resilience against adversarial inputs frequently comes at the cost of a significant drop in performance on clean, unperturbed data~\cite{zhang2020revisiting, wang2021measure}. This robustness-accuracy challenge not only poses a significant barrier to the reliable deployment of PLMs in safety-critical applications but also highlights a fundamental gap in our understanding of their internal workings.

\begin{figure}[t]
\centering
\includegraphics[width=0.95\columnwidth]{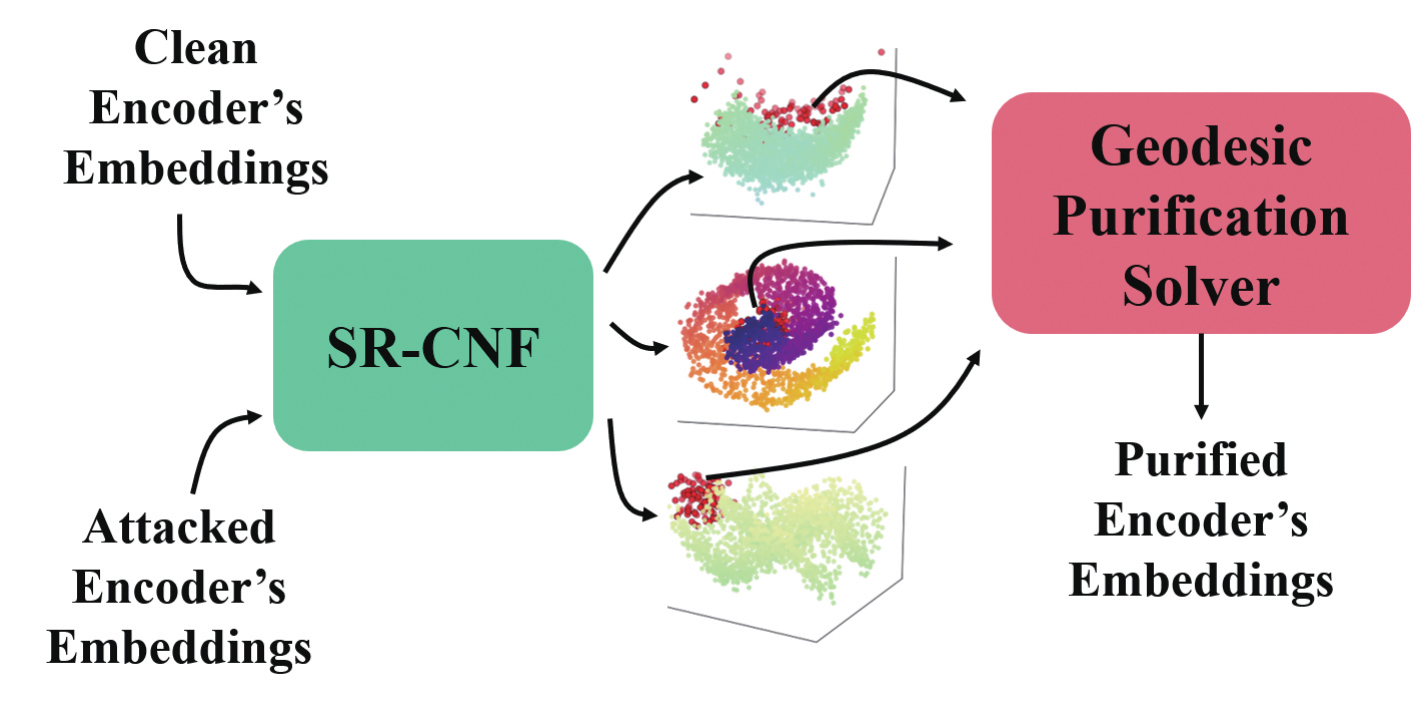} 
\caption{An overview of our proposed Manifold-Correcting Causal Flow ($MC^{2}F$) framework. The system first employs a Stratified Riemannian Continuous Normalizing Flow (SR-CNF) to model the manifold of clean data (colored points). It then identifies out-of-distribution (OOD) or adversarial samples (red points) that lie off the manifold. The Purification Solver module subsequently corrects these OOD embeddings by projecting them back onto the clean manifold along the shortest path (geodesic), a process visualized by the black arrows. In-distribution embeddings are passed through without modification.}
\label{fig1}
\end{figure}

To overcome this challenge, we move beyond heuristic defenses and investigate the geometric properties of the model's embedding space. We begin with a critical empirical observation: in the high-dimensional embedding space of a BERT-style encoder, representations of clean text are geometrically separable from their adversarial counterparts generated by prevalent attacks (e.g., TextFooler, BERT-Attack)~\cite{subhash2023universal,reif2019visualizing}. This suggests that adversarial perturbations induce a measurable geometric and distributional shift, rather than creating indistinguishable overlaps. Consequently, the defense problem can be reframed: instead of a brute-force training task, it becomes a more nuanced geometric challenge of identifying and correcting embeddings that have been perturbed off the "clean" data distribution.

Building on this insight and inspired by the manifold hypothesis—which posits that high-dimensional data lies on a lower-dimensional manifold—we introduce the Manifold-Correcting Causal Flow ($MC^{2}F$). It is a two-module defense system that operates directly on sentence embeddings, embodying a "detect-and-correct" method to purify adversarial inputs ~\cite{li2023text, moraffah2024adversarial}.

The first module, a Stratified Riemannian Continuous Normalizing Flow (SR-CNF), is responsible for detection. It learns a probabilistic model of the clean data manifold, leveraging a dynamically learned Riemannian metric to perform highly accurate density estimation. This allows it to effectively distinguish in-distribution ("clean") embeddings from out-of-distribution (OOD) adversarial samples based on a learned likelihood threshold.

For any embedding flagged as adversarial, the second module, a Geodesic Purification Solver, performs the correction. We formalize this process as an optimization problem: finding the shortest path (geodesic) from the anomalous point back to the learned manifold. The solver iteratively finds this path to produce a purified, semantically robust embedding, effectively neutralizing the adversarial perturbation.

The contributions of this paper are summarized as follows:

\begin{itemize}
    \item We provide strong empirical evidence that the embeddings of clean and adversarial text are geometrically and distributionally separable in the representation space of Pre-trained Language Models. This insight allows us to reframe adversarial defense from a brute-force training problem to a geometric purification task.

    \item We propose the Manifold-Correcting Causal Flow (${MC^2F}$), a novel and principled framework that operationalizes our geometric insight. $MC^2F$ is the first system to integrate a Stratified Riemannian Continuous Normalizing Flow for out-of-distribution detection with a Geodesic Purification Solver for correction.

    \item Through extensive experiments across three datasets, two downstream tasks, and multiple attack methods, we demonstrate that $MC^2F$ effectively resolves the long-standing robustness-accuracy trade-off. It achieves state-of-the-art adversarial robustness while fully preserving---and in some cases, even improving---performance on clean data.
\end{itemize}

\section{Related Works}

Our work is positioned at the intersection of adversarial defense in NLP and the geometric analysis of embedding spaces. We review the relevant literature in these areas.

\subsection{Adversarial Attacks in Text Classification}
The vulnerability of PLMs has been extensively demonstrated by a variety of adversarial attack methods. These methods typically generate perturbations by replacing, inserting, or deleting characters or words under certain constraints to maintain semantic similarity and fluency. Prominent word-level attack algorithms include TextFooler~\cite{jin2020bert}, BAE~\cite{garg2020bae}, and TextBugger~\cite{li2018textbugger}, which greedily replace words with synonyms that maximize the victim model's prediction error. More sophisticated methods, such as BERT-Attack~\cite{li2020bert}, leverage the victim model itself to find more effective substitutes. These attacks serve as standard benchmarks for evaluating the robustness of defense systems, including our own.

\subsection{Adversarial Defense Strategies}

The dominant paradigm, adversarial training (AT), augments data with adversarial examples~\cite{zhu2020freelb, madry2018towards} yet suffers from high computational costs and accuracy trade-offs~\cite{zhang2020revisiting, wang2021measure}. Despite improvements, an “illusion of robustness” often persists due to gradient masking~\cite{gao2023dsrm, wang2024new, raina2024extreme}. Alternatively, purification methods have evolved toward continuous embedding denoising~\cite{zheng2024subspace, asl2024robustsentembed}, with recent works like DAD balancing clean and robust accuracy via MMD-based detection~\citep{zhang2025ddad}. A complementary geometric perspective relies on the manifold hypothesis~\cite{fefferman2016testing}, projecting off-manifold adversarial points back to the data manifold~\cite{minh2022textual, yang2024adversarial}. Motivated by evidence that off-manifold samples drive robustness while on-manifold data supports generalization~\cite{altinisik2024explaining, dang2025meria}, we pursue a formal manifold-based NLP purification framework.

\subsection{Flow-based Out-of-Distribution Detection}

Adversarial detection can be cast as an out-of-distribution (OOD) task, where clean and adversarial samples represent in- and out-distribution data, respectively. Generative models like Normalizing Flows (NFs)~\cite{rezende2015variational} and Continuous NFs (CNFs)~\cite{chen2018neural} excel at this by assigning low likelihoods to OOD samples~\cite{nalisnick2019deep}. To overcome limitations in modeling complex manifolds, modern approaches enhance performance by applying flows in feature spaces~\cite{cook2024feature} or integrating Riemannian metrics for precise geometric density estimation~\cite{diepeveen2024score}.

\begin{figure*}[!ht]
\centering

\begin{subfigure}[b]{0.46\textwidth}
    \centering
    \includegraphics[width=\textwidth]{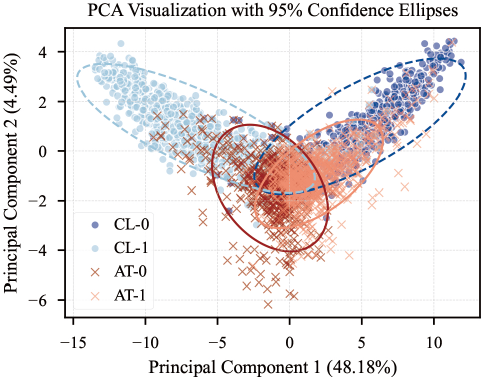}
    \caption{}
    \label{fig:pca}
\end{subfigure}
\hfill
\begin{subfigure}[b]{0.46\textwidth}
    \centering
    \includegraphics[width=\textwidth]{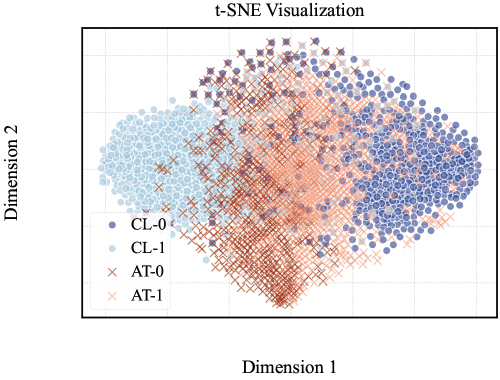}
    \caption{}
    \label{fig:tsne}
\end{subfigure}

\vspace{4pt}

\begin{subfigure}[b]{0.46\textwidth}
    \centering
    \includegraphics[width=\textwidth]{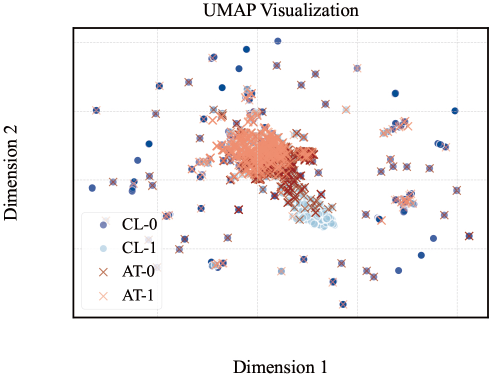}
    \caption{}
    \label{fig:umap}
\end{subfigure}
\hfill
\begin{subfigure}[b]{0.46\textwidth}
    \centering
    \includegraphics[width=\textwidth]{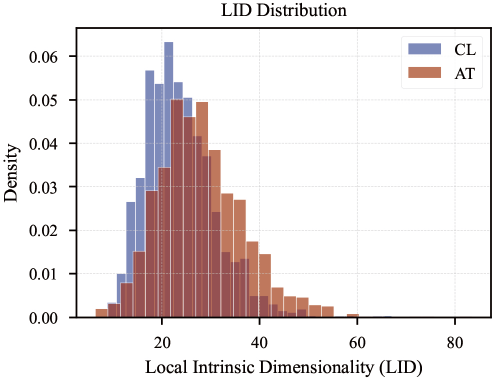}
    \caption{}
    \label{fig:lid_hist}
\end{subfigure}

\caption{Geometric and distributional analysis of clean and attacked SST-2 test set embeddings. (a-c) 2D visualizations using PCA, t-SNE, and UMAP show that adversarial embeddings form clusters visibly distinct from clean embeddings. (d) The density distribution of Local Intrinsic Dimensionality (LID) values shows that attacked embeddings are visibly shifted towards higher LID values. CL-0/1: Clean samples of class 0/1; AT-0/1: Attacked samples of class 0/1.}
\label{fig:visualizations_and_lid}
\end{figure*}

\section{Preliminary Study}
\label{sec:preliminary_study}

Our defense strategy is predicated on the hypothesis that adversarial perturbations displace text embeddings from a "clean" data manifold into distinct, separable regions of the embedding space. This section presents a rigorous empirical investigation to validate this hypothesis. We analyze the geometric and distributional properties of encoder's embeddings from a BERT-base model ~\cite{devlin2019bert} fine-tuned on the SST-2 dataset ~\cite{socher2013recursive}, comparing clean test set examples against their adversarial counterparts generated by TextFooler~\cite{jin2020bert}.

\subsection{Problem Formulation}
Let $f_\theta: \mathcal{X} \to \mathcal{Z}$ be a text encoder, such as BERT, mapping an input text $x \in \mathcal{X}$ to an embedding $z \in \mathcal{Z} \subseteq \mathbb{R}^d$. Let $P_{clean}$ denote the probability distribution of embeddings $z = f_\theta(x)$ derived from a clean data distribution. An adversarial attack generates a perturbed input $x_{adv}$ from $x$, resulting in an embedding $z_{adv} = f_\theta(x_{adv})$. Let $P_{adv}$ be the distribution of these adversarial embeddings. Our central premise is that $P_{clean}$ and $P_{adv}$ are significantly different, and more specifically, that their supports lie on separable, low-dimensional manifolds, $\mathcal{M}_{clean}$ and $\mathcal{M}_{adv}$, respectively.

\subsection{Visual Evidence of Manifold Separation}
To first gain a visual intuition, we project the high-dimensional embeddings into a 2D space using three different techniques: Principal Component Analysis (PCA), t-Distributed Stochastic Neighbor Embedding (t-SNE)~\cite{maaten2008visualizing}, and Uniform Manifold Approximation and Projection (UMAP)~\cite{mcinnes2018umap}. As shown in Figure~\ref{fig:visualizations_and_lid}(a-c), all three methods reveal a clear separation between the clean (blue circles) and attacked (red crosses) embeddings. While PCA, a linear projection method, shows a noticeable distributional shift, the non-linear, manifold-aware methods t-SNE and UMAP illustrate a more striking separation. In the UMAP projection (Figure~\ref{fig:umap}), the clean and attacked points form well-defined, dense clusters with minimal overlap. This strong visual evidence suggests that clean and adversarial embeddings occupy distinct regions in the latent space, consistent with our manifold separation hypothesis.

\begin{table*}[t]
\centering
\setlength{\tabcolsep}{4pt} 
\begin{tabular}{@{}lccc@{}}
\toprule
\textbf{Comparison} & \textbf{MMD (Gaussian Kernel) $\downarrow$} & \textbf{JSD $\downarrow$} & \textbf{Wasserstein Distance $\downarrow$} \\ \midrule
Clean In-Distribution (Class 0 Half 1 vs. Half 2) & 0.00068 & 0.00008 & 19.66 \\ \midrule
All Clean vs. All Attacked & \textbf{0.00254} & \textbf{0.00096} & \textbf{30.76} \\
Class 0: Clean vs. Attacked & \textbf{0.00353} & \textbf{0.00100} & \textbf{88.86} \\
Class 1: Clean vs. Attacked & \textbf{0.00183} & \textbf{0.00940} & \textbf{56.31} \\ \bottomrule
\end{tabular}
\caption{Statistical distances between embedding distributions. The distances between clean and attacked distributions are consistently and significantly larger than the in-distribution baseline, providing quantitative proof of a distributional shift caused by adversarial attacks.}
\label{tab:dist_metrics}
\end{table*}

\subsection{Quantitative Evidence of Distributional Shift}
To move beyond visual intuition, we quantify the dissimilarity between the clean and adversarial embedding distributions using three standard metrics: Maximum Mean Discrepancy (MMD) with a Gaussian kernel ~\cite{gretton2012kernel}, Jensen-Shannon Divergence (JSD)~\cite{lin2002divergence}, and the Wasserstein Distance. Table~\ref{tab:dist_metrics} reports these distances. For a baseline, we first measure the distance between two random halves of the clean embeddings ("Clean In-Distribution"), which is expectedly low. In stark contrast, the distances between the full set of clean and attacked embeddings ("All Clean vs. All Attacked") are substantially larger across all three metrics. This pattern holds true when examining each class individually. These results provide strong quantitative evidence that adversarial attacks induce a statistically significant distributional shift, confirming that $P_{clean}$ and $P_{adv}$ are indeed different distributions.

\begin{table}[b]
\centering
\setlength{\tabcolsep}{2.5pt}
\begin{tabular}{@{}lccc@{}}
\toprule
\textbf{Comparison} & \textbf{MLID (CL)} & \textbf{MLID (AT)} & \textbf{p-value} \\ \midrule
All Classes & 23.74 & 28.20 & $8.99 \times 10^{-43}$ \\
Class 0 & 23.67 & 25.66 & $2.36 \times 10^{-5}$ \\
Class 1 & 21.75 & 26.90 & $5.50 \times 10^{-20}$ \\ \bottomrule
\end{tabular}

\caption{Mean LID (MLID) values and Welch's t-test results. The significantly higher mean LID for attacked embeddings and the extremely low p-value suggest that adversarial perturbations move points to regions of higher geometric complexity. CL denotes clean embeddings and AT denotes attacked embeddings.}
\label{tab:lid_stats}
\end{table}

\subsection{Analysis of Local Intrinsic Dimensionality}
Finally, we investigate if the geometric structure of the manifolds themselves differs. We use Local Intrinsic Dimensionality (LID)~\cite{houle2017local}, which estimates the dimensionality of the data manifold in the local neighborhood of a data point. A higher LID suggests greater local complexity.

Figure~\ref{fig:lid_hist} shows the density distributions of LID values for both clean and attacked embeddings. A clear rightward shift is visible for the attacked distribution. This observation is statistically validated in Table~\ref{tab:lid_stats}. The mean LID of attacked embeddings (28.20) is significantly higher than that of clean embeddings (23.74). A Welch's t-test confirms this difference is highly statistically significant, with a p-value approaching zero ($p \approx 10^{-43}$). This implies that adversarial perturbations do not just move embeddings to a different location, but systematically push them into regions of the embedding space that are geometrically more complex.

\subsection{Foundational Hypotheses}

Based on this comprehensive empirical evidence and recent advances in the study of embeddings from language models ~\cite{li2025unraveling, mamou2020emergence}, we establish two foundational hypotheses that motivate our proposed method:

\begin{enumerate}
    \item \textbf{Manifold Separability:} The embeddings of clean text and their adversarial counterparts lie on statistically distinct and geometrically separable manifolds within the encoder's representation space.
    \item \textbf{Stratified Manifold Structure:} The embedding space is not a single, uniform manifold but a stratified space composed of sub-manifolds with varying local geometric properties, such as different intrinsic dimensionalities. Adversarial examples tend to occupy regions of higher intrinsic dimensionality than clean examples.
\end{enumerate}
These hypotheses suggest that an effective defense can be built by first learning the manifold of clean data and then correcting adversarial points by projecting them back onto this learned manifold, which is the core idea behind our proposed ${MC^2F}$ framework.

\section{Method}
\label{sec:method}
Building upon our foundational hypotheses, we introduce the Manifold-Correcting Causal Flow ($MC^2F$). This framework is designed not only to detect adversarial samples by modeling the complex geometry of the clean data manifold, $\mathcal{M}_{clean}$, but also to purify them by projecting them back onto this manifold. $MC^2F$ comprises two core modules: a Stratified Riemannian Continuous Normalizing Flow (SR-CNF) for detection and a Geodesic Purification Solver for correction. The $MC^2F$ framework is summarized in Algorithm~\ref{alg:mc2f}.

\begin{algorithm}[tb]
\caption{$MC^2F$: Manifold-Correcting Causal Flow}
\label{alg:mc2f}
\begin{algorithmic}[1]
\STATE \textbf{Require:} Clean data embeddings $Z_{clean}$, downstream classifier $C$, hyperparameters $\lambda_{topo}, \lambda_{causal}$.
\STATE Initialize parameters $\Theta_{flow}$ for the SR-CNF map $f$ and $\Theta_{metric}$ for the MoE metric $G(z)$.

\vspace{0.5em}
\STATE \textbf{// Training Phase}
\FOR{each training batch $Z_b \subset Z_{clean}$}
    \STATE Get adversarial counterparts $Z_{adv}$ from $Z_b$.
    \STATE Purify $Z_{adv}$ to get $Z_{corr}$ using the Geodesic Purification Solver (Eq.~\ref{eq:geodesic_energy}) with the current metric $G(z)$.
    \STATE Compute $\mathcal{L}_{NLL}$ on $Z_b$ (Eq.~\ref{eq:log_likelihood}).
    \STATE // Compute latent counterparts by inverting the flow
    \STATE Compute $\mathcal{L}_{topo}$ on $Z_b$ and $f^{-1}(Z_b)$ (Eq.~\ref{eq:topo_loss}).
    \STATE Compute $\mathcal{L}_{causal}$ between $C(Z_b)$ and $C(Z_{corr})$.
    \STATE Compute total loss $\mathcal{L}_{total} = \mathcal{L}_{NLL} + \lambda_{topo}\mathcal{L}_{topo} + \lambda_{causal}\mathcal{L}_{causal}$.
    \STATE Update $\Theta_{flow}$ and $\Theta_{metric}$ via gradient descent.
\ENDFOR
\STATE Determine likelihood threshold $\tau$ on a validation set.

\vspace{0.5em}
\STATE \textbf{// Inference Phase}
\FOR{each input embedding $z_{in}$ to be classified}
    \STATE Calculate $\ell = \log p(z_{in})$ using SR-CNF.
    \IF{$\ell < \tau$}
        \STATE $z_{out} \leftarrow \text{Geodesic Purification Solver}(z_{in})$
    \ELSE
        \STATE $z_{out} \leftarrow z_{in}$
    \ENDIF
    \STATE Obtain final prediction from $C(z_{out})$.
\ENDFOR
\end{algorithmic}
\end{algorithm}

\subsection{Stratified Riemannian CNF for Detection}
To accurately model the probability density of clean embeddings, $p_{clean}(z)$, we must account for the stratified and non-Euclidean nature of the embedding space. We achieve this by defining a Continuous Normalizing Flow (CNF) on a learnable, data-dependent Riemannian manifold.

\subsubsection{Learning a Stratified Riemannian Geometry}
Instead of assuming a fixed geometry, we learn a Riemannian metric tensor $G(z)$, a positive definite matrix that defines the local geometry at each point $z \in \mathcal{Z}$. To capture the stratified structure observed in our preliminary study, we parameterize $G(z)$ using a Mixture-of-Experts (MoE) network~\cite{jacobs1991adaptive, fedus2022switch}. This network consists of:
\begin{itemize}
    \item A gating network $g_\phi(z)$ that takes an embedding $z$ and outputs a set of weights $\{\alpha_k(z)\}_{k=1}^K$, where $K$ is the number of experts and $\sum_k \alpha_k(z) = 1$.
    \item $K$ expert networks $\{E_{\psi_k}(z)\}_{k=1}^K$, where each expert is a neural network that outputs a matrix, specializing in the local geometry of a specific stratum.
\end{itemize}
The final metric tensor at point $z$ is a weighted combination of the expert outputs:
\begin{equation}
    G(z) = \sum_{k=1}^{K} \alpha_k(z) E_{\psi_k}(z).
    \label{eq:expo}
\end{equation}
To ensure $G(z)$ is always positive definite, each expert $E_{\psi_k}(z)$ is constructed as $L_k(z)L_k(z)^T + \epsilon I$, where $L_k(z)$ is the output of the expert network (a lower-triangular matrix) and $\epsilon$ is a small positive constant for numerical stability. This MoE structure allows the model to adaptively learn a complex, piece-wise smooth geometry across different semantic regions of the embedding space.

\subsubsection{Riemannian Continuous Normalizing Flow}
With the learned Riemannian manifold $(\mathcal{M}, G)$, we define a flow as the solution to the ordinary differential equation (ODE) governed by a time-varying vector field $v$:
\begin{equation}
    \frac{dz(t)}{dt} = v(z(t), t), \quad z(t_0) \sim \mathcal{N}(0, I),
\end{equation}
where the dynamics unfold on the manifold $\mathcal{M}$. The change of variable formula for a CNF on a Riemannian manifold relates the change in log-probability to the Riemannian divergence of the vector field:
\begin{equation}
    \frac{d \log p(z(t))}{dt} = -\text{div}_G(v(z(t), t)).
\end{equation}

The total log-likelihood of a data point $z_{in}$ is then obtained by integrating this quantity along the forward trajectory from its latent representation $z_0$ at $t=t_0$ to the data point $z_{in}$ at $t=t_1$:

\begin{equation}
\label{eq:log_likelihood}
    \log p(z_{in}) = \log p_{\mathcal{N}}(z_0) - \int_{t_0}^{t_1} \text{div}_G(v(z(t), t)) dt.
\end{equation}
\textbf{Detection Mechanism:} An input embedding $z_{in}$ is classified as adversarial if its log-likelihood falls below a threshold $\tau$, i.e., $\log p(z_{in}) < \tau$. The threshold $\tau$ is determined on a validation set to achieve a desired false positive rate.

\begin{table*}[t]
\centering
\begin{tabular}{@{}ll|c|cc|cc|cc@{}}
\toprule
\multirow{2}{*}{\textbf{Dataset}} & \multirow{2}{*}{\textbf{Method}} & \multirow{2}{*}{\textbf{Clean\%}} & \multicolumn{2}{c|}{\textbf{BERT-Attack}} & \multicolumn{2}{c|}{\textbf{TextFooler}} & \multicolumn{2}{c}{\textbf{TextBugger}} \\
\cmidrule(l){4-9} 
 &  &  & \textbf{Aua\%} & \textbf{\#Query} & \textbf{Aua\%} & \textbf{\#Query} & \textbf{Aua\%} & \textbf{\#Query} \\ \midrule
\multirow{5}{*}{SST-2} & Fine-tune & 92.71 & 3.83 & 106.4 & 6.10 & 90.5 & 28.70 & 46.0 \\
 & FreeLB & 92.01 & 23.88 & 174.7 & 29.40 & 132.6 & 49.70 & 53.8 \\
 & WLRE & 92.11 & 29.80 & 185.4 & 32.80 & 138.4 & 50.10 & 56.4 \\
 & SD & 91.36 & 36.46 & 201.2 & 46.30 & 167.3 & 54.50 & 62.3 \\
 & $\mathbf{MC^2F}$ & \textbf{92.71} & \textbf{40.05} & \textbf{289.4} & \textbf{52.60} & \textbf{184.2} & \textbf{61.50} & \textbf{98.4} \\ \midrule
\multirow{5}{*}{AGNews} & Fine-tune & 94.68 & 4.09 & 412.9 & 14.70 & 306.4 & 40.00 & 166.2 \\
 & FreeLB & 94.99 & 19.90 & 581.8 & 33.20 & 396.0 & 52.90 & 201.1 \\
 & WLRE & 94.05 & 28.60 & 657.1 & 32.60 & 368.1 & 53.40 & 208.0 \\
 & SD & 93.81 & 38.60 & 744.1 & 49.30 & 488.1 & 60.10 & 219.7 \\
 & $\mathbf{MC^2F}$ & \textbf{95.13} & \textbf{45.30} & \textbf{892.5} & \textbf{53.80} & \textbf{561.4} & \textbf{64.30} & \textbf{299.0} \\ \midrule
\multirow{5}{*}{YELP} & Fine-tune & 95.19 & 5.40 & 116.2 & 5.20 & 105.4 & 29.60 & 52.6 \\
 & FreeLB & 95.11 & 28.24 & 184.3 & 28.30 & 143.5 & 50.00 & 68.4 \\
 & WLRE & 94.86 & 31.46 & 208.2 & 31.50 & 155.0 & 50.00 & 69.1 \\
 & SD & 93.45 & 39.61 & 320.7 & 47.80 & 187.6 & 55.10 & 100.2 \\
 & $\mathbf{MC^2F}$ & \textbf{95.26} & \textbf{48.50} & \textbf{586.4} & \textbf{54.00} & \textbf{214.3} & \textbf{63.20} & \textbf{112.4} \\ \bottomrule
\end{tabular}
\caption{Main results on adversarial robustness evaluation. We report accuracy on the clean test set (Clean\%) and accuracy under attack (Aua\%), along with the average number of queries per successful attack (\#Query). Higher is better for all metrics. Our method achieves the best robustness across all datasets and attacks while maintaining high clean accuracy.}
\label{tab:main_results}
\end{table*}

\subsection{Geodesic Purification Solver}
When an embedding $z_{adv}$ is detected as adversarial, the correction module is activated. Its goal is to find a purified point $z_{corr}$ on the clean manifold $\mathcal{M}_{clean}$ that is "closest" to $z_{adv}$. In the context of our learned Riemannian geometry, this corresponds to finding the orthogonal projection of $z_{adv}$ onto $\mathcal{M}_{clean}$. This projection point lies at the end of a geodesic (the shortest path on the manifold) starting at $z_{adv}$ and ending on $\mathcal{M}_{clean}$.

We formulate this as an optimization problem to find the path $\gamma(t): [0, 1] \to \mathbb{R}^d$ that minimizes the path energy functional:
\begin{equation}
\label{eq:geodesic_energy}
    \mathcal{L}[\gamma] = \int_{0}^{1} \langle \gamma'(t), \gamma'(t) \rangle_{G(\gamma(t))} dt,
\end{equation}
subject to the boundary conditions $\gamma(0) = z_{adv}$ and $\gamma(1) = z_{corr} \in \mathcal{M}_{clean}$. The optimal path $\gamma^*$ is the geodesic, and its endpoint $\gamma^*(1)$ is our purified embedding $z_{corr}$. 

This optimization problem is solved iteratively. We discretize the path $\gamma(t)$ and use gradient descent on its waypoints to minimize the energy functional $\mathcal{L}[\gamma]$ (Eq.~\ref{eq:geodesic_energy}), with gradients computed respecting the learned metric $G(z)$ (Eq.~~\ref{eq:expo}). The constraint $\gamma(1) \in \mathcal{M}_{clean}$ is operationalized by enforcing $\log p(z_{corr}) \ge \tau$ (where $\tau$ is the detection threshold) via a soft penalty term in the minimization objective.

\subsection{Geometric and Topological Training Paradigm}
To effectively train the $MC^2F$ framework, we employ a multi-objective loss function that optimizes for density estimation, topological integrity, and semantic consistency simultaneously. The total loss is:
\begin{equation}
    \mathcal{L}_{total} = \mathcal{L}_{NLL} + \lambda_{topo}\mathcal{L}_{topo} + \lambda_{causal}\mathcal{L}_{causal},
\end{equation}
where $\lambda_{topo}$ and $\lambda_{causal}$ are balancing hyperparameters.

\subsubsection{Density Estimation Loss ($\mathcal{L}_{NLL}$)}
This is the standard negative log-likelihood loss for training normalizing flows. It drives the model to learn the distribution of clean data:
\begin{equation}
    \mathcal{L}_{NLL} = -\mathbb{E}_{z \sim P_{clean}}[\log p(z)],
\end{equation}
where $\log p(z)$ is computed using our SR-CNF (Eq.~\ref{eq:log_likelihood}).

\subsubsection{Topological Regularization ($\mathcal{L}_{topo}$)}
To ensure the learned flow (SR-CNF) $f$ preserves the global shape of the data manifold and avoids topological tearing, we introduce a loss based on differentiable persistent homology. We compute the persistence diagrams (PD), topological summaries of data, for a batch of clean embeddings $Z_{b}$ and their latent counterparts $f^{-1}(Z_{b})$. The loss is the Wasserstein distance between these diagrams:
\begin{equation}
    \mathcal{L}_{topo} = W_p(\text{PD}(Z_{b}), \text{PD}(f^{-1}(Z_{b}))).
    \label{eq:topo_loss}
\end{equation}
Minimizing this loss encourages the flow to be a homeomorphism, preserving the topological features of the data manifold.

\subsubsection{Causal \& Semantic Regularization ($\mathcal{L}_{causal}$)}
The purification process $z_{adv} \to z_{corr}$ must be semantically consistent. We frame this from a causal perspective, where purification is an intervention to remove the confounding effect of the adversarial perturbation. To enforce this, we use a loss based on information geometry ~\cite{volpi2021changing, kim2022fisher}. For a clean embedding $z_{clean}$ and its corresponding purified version $z_{corr}$ (obtained by first attacking $z_{clean}$ and then purifying it), we pass both through a fixed, pre-trained downstream classifier $C(\cdot)$ to get two softmax probability distributions, $p_{clean}$ and $p_{corr}$. The causal loss is the Fisher-Rao distance between them:
\begin{equation}
    \mathcal{L}_{causal} = 2 \arccos\left(\sum_{i} \sqrt{p_{clean,i} \cdot p_{corr,i}}\right).
\end{equation}
This loss directly forces the purified embedding to be semantically indistinguishable from the original clean embedding from the perspective of the downstream task.

\section{Experiments}
This section presents a comprehensive set of experiments designed to rigorously evaluate the effectiveness of our proposed $MC^2F$ framework. 

\subsection{Experimental Setup}

\subsubsection{Datasets.}
We evaluate our method on three benchmark datasets: SST-2~\cite{socher2013recursive} for binary sentiment analysis on short text; AGNews~\cite{jacovi2018understanding} for four-class news topic classification; and YELP~\cite{asghar2016yelp}, a more complex binary sentiment task involving longer reviews.

\subsubsection{Attack Methods.}
We assess robustness using the TextAttack toolkit~\cite{morris2020textattack} with three established attacks: BERT-Attack~\cite{li2020bert}, which leverages BERT for context-aware substitutions; TextFooler~\cite{jin2020bert}, a greedy strategy using synonym replacement; and TextBugger~\cite{li2018textbugger}, which employs both word- and character-level perturbations.

\subsubsection{Baselines.}
We benchmark $MC^2F$ against a standard fine-tuned model and three competitive defense baselines:
\begin{itemize}
    \item Fine-tune ~\cite{devlin2019bert}: A standard BERT-base model fine-tuned for each downstream task, representing a non-robust baseline.
    \item FreeLB~\cite{zhu2020freelb}: An enhanced adversarial training method that adds adversarial perturbations to word embeddings within a bounded region to improve generalization and robustness.
    \item WLRE~\cite{huang2022word}: A defense framework that enhances robustness by introducing its own perturbations to the input during training, forcing the model to learn more invariant features.
    \item SD (Subspace Defense)~\cite{zheng2024subspace}: A state-of-the-art defense that projects feature representations onto a learned "clean" subspace to discard adversarial perturbations.
\end{itemize}

\subsubsection{Evaluation Metrics.}
We evaluate all methods based on three metrics:
\begin{itemize}
    \item Clean\%: The classification accuracy on the original, unperturbed test set. This metric is crucial for assessing whether a defense method compromises performance on normal data.
    \item Aua\%: The model's accuracy on the adversarially attacked test set. This is the primary metric for evaluating adversarial robustness.
    \item \#Query: The average number of model queries required by the attacker to successfully generate an adversarial example. A higher query count indicates a more resilient model.
\end{itemize}

\begin{table}[t]
\centering
\begin{tabular}{@{}l|c|cc@{}}
\toprule
\multirow{2}{*}{\textbf{Method}} & \multirow{2}{*}{\textbf{Clean\%}} & \multicolumn{2}{c}{\textbf{TextFooler}} \\ \cmidrule(l){3-4} 
 &  & \textbf{Aua\%} & \textbf{\#Query} \\ \midrule
Fine-tune & 94.68 & 14.7 & 306.4 \\
$\mathbf{MC^2F}$ \textbf{(Full)} & \textbf{95.13} & \textbf{53.8} & \textbf{561.4} \\ \midrule
w/o $\mathcal{L}_{NLL}$ & 93.22 & 32.6 & 366.7 \\
w/o $\mathcal{L}_{topo}$ & 93.41 & 32.9 & 375.4 \\
w/o $\mathcal{L}_{causal}$ & 94.76 & 48.6 & 479.1 \\ \bottomrule
\end{tabular}
\caption{Ablation study on the AGNews dataset against the TextFooler attack. The results demonstrate that each component of our multi-objective loss function is essential for achieving optimal performance.}
\label{tab:ablation}
\end{table}

\subsubsection{Implementation Details.}
All experiments use BERT-base-uncased as the backbone encoder. For our $MC^2F$ framework, we set the key hyperparameters based on validation performance: the likelihood threshold for detection $\tau=0.3$, the weight for the topological loss $\lambda_{topo}=0.8$, and the weight for the causal loss $\lambda_{causal}=0.05$.

\subsection{Main Results on Robustness Evaluation}
The primary experimental results are summarized in Table~\ref{tab:main_results}. The comparison across three datasets and three attack methods reveals several key findings.

First and most importantly, $MC^2F$ consistently and substantially outperforms all baseline methods in robust accuracy (Aua\%) across all evaluated datasets and attack scenarios. For instance, under the strong BERT-Attack on the AGNews dataset, $MC^2F$ achieves an Aua of 45.3\%, a significant improvement over the next best baseline, SD, which scores 38.6\%. This pattern of superior performance holds for TextFooler and TextBugger attacks as well, demonstrating the broad effectiveness of our geometric purification approach.

Second, this significant gain in robustness does not come at the cost of performance on clean data. The Clean\% of $MC^2F$ is on par with, and in the cases of AGNews and YELP, even slightly higher than the standard Fine-tune model. This successfully addresses the critical robustness-accuracy trade-off that plagues many existing defense methods like SD, which shows a noticeable drop in clean accuracy.

Third, $MC^2F$ demonstrates heightened resilience as measured by the number of queries (\#Query). Attackers consistently require more queries to find successful adversarial examples against our method compared to all baselines. On YELP against BERT-Attack, for example, an attacker needs an average of 586.4 queries for $MC^2F$, compared to 320.7 for SD. This indicates that our defense makes the model's decision boundaries more stable and harder to exploit.

\subsection{Ablation Study}
To isolate and verify the contribution of each key component in our proposed training paradigm, we conduct an ablation study on the AGNews dataset under the TextFooler attack. We systematically remove each of the three loss terms from our objective function. The results are presented in Table~\ref{tab:ablation}.

The analysis reveals that each component is integral to the framework's success.
Removing the negative log-likelihood loss ($\mathcal{L}_{NLL}$) results in a catastrophic drop in both clean and robust accuracy. This is expected, as $\mathcal{L}_{NLL}$ is fundamental for learning the clean data distribution, which is the basis for our detection module.

Excluding the topological regularization term ($\mathcal{L}_{topo}$) causes the most significant drop in robust accuracy among the regularization terms, with Aua\% falling from 53.8\% to 32.9\%. This result strongly supports our hypothesis that preserving the global topological structure of the data manifold is critical for preventing the model from learning a brittle representation that is easily exploited by adversaries.

Removing the causal and semantic loss ($\mathcal{L}_{causal}$) also leads to a notable decrease in robustness, with Aua\% dropping to 48.6\%. This demonstrates the importance of explicitly guiding the purification process to be semantically consistent with the original clean input from the perspective of the downstream classifier.

\section{Conclusion}
In this paper, we addressed the persistent trade-off between robustness and accuracy in text classification by proposing a novel defense rooted in the geometric properties of the embedding space. Our framework, $MC^2F$, operationalizes the manifold hypothesis by first learning a stratified Riemannian geometry of clean data with a continuous normalizing flow, and then purifying detected adversarial samples via geodesic projection. The experiments demonstrate that $MC^2F$ not only sets a new state-of-the-art in adversarial robustness against multiple attacks but, crucially, does so without degrading performance on clean data. This work validates the power of a geometric, detect-and-correct approach, paving the way for developing more principled and robust NLP systems that resolve the long-standing robustness-accuracy dilemma.

\bibliography{aaai2026}

\end{document}